\def\year{2021}\relax
\documentclass[letterpaper]{article} 
\usepackage{aaai21}  
\usepackage{times}  
\usepackage{helvet} 
\usepackage{courier}  
\usepackage[hyphens]{url}  
\usepackage{graphicx} 
\usepackage{natbib}  
\usepackage{caption} 
\usepackage{booktabs}
\usepackage{subcaption}
\usepackage{multirow}
\usepackage{amssymb}
\usepackage[linesnumbered, algoruled]{algorithm2e}
\usepackage{amsmath,amssymb}
\usepackage[pagebackref=true,breaklinks=true,letterpaper=true,colorlinks,bookmarks=false]{hyperref}       

\newcommand{\etal}{\textit{et al}. }
\newcommand{\ie}{\textit{i}.\textit{e}., }

\title{SCNet: Training Inference Sample Consistency for Instance Segmentation}
\author {
    Thang Vu, \qquad
    Haeyong Kang, \qquad
    Chang D. Yoo\\
}
\affiliations {
    Department of Electrical Engineering \\
    Korea Advanced Institute of Science and Technology \\
    \texttt{\{thangvubk,haeyong.kang,cd\_yoo\}@kaist.ac.kr}
}

\begin{document}
	\maketitle

	\begin{abstract}
	Cascaded architectures have brought significant performance improvement in object detection and instance segmentation. However, there are lingering issues regarding the disparity in the Intersection-over-Union (IoU) distribution of the samples between training and inference. This disparity can potentially exacerbate detection accuracy. This paper proposes an architecture referred to as Sample Consistency Network (SCNet) to ensure that the IoU distribution of the samples at training time is close to that at inference time. Furthermore, SCNet incorporates feature relay and utilizes global contextual information to further reinforce the reciprocal relationships among classifying, detecting, and segmenting sub-tasks. Extensive experiments on the standard COCO dataset reveal the effectiveness of the proposed method over multiple evaluation metrics, including box AP, mask AP, and inference speed. In particular, while running 38\% faster, the proposed SCNet improves the AP of the box and mask predictions by respectively 1.3 and 2.3 points compared to the strong Cascade Mask R-CNN baseline. Code is available at \url{https://github.com/thangvubk/SCNet}.
\end{abstract}

\begin{figure*}[!t]
	\centering
	\begin{subfigure}{0.32\textwidth}
		\centering
		\includegraphics[width=1\textwidth, height=0.7\textwidth]{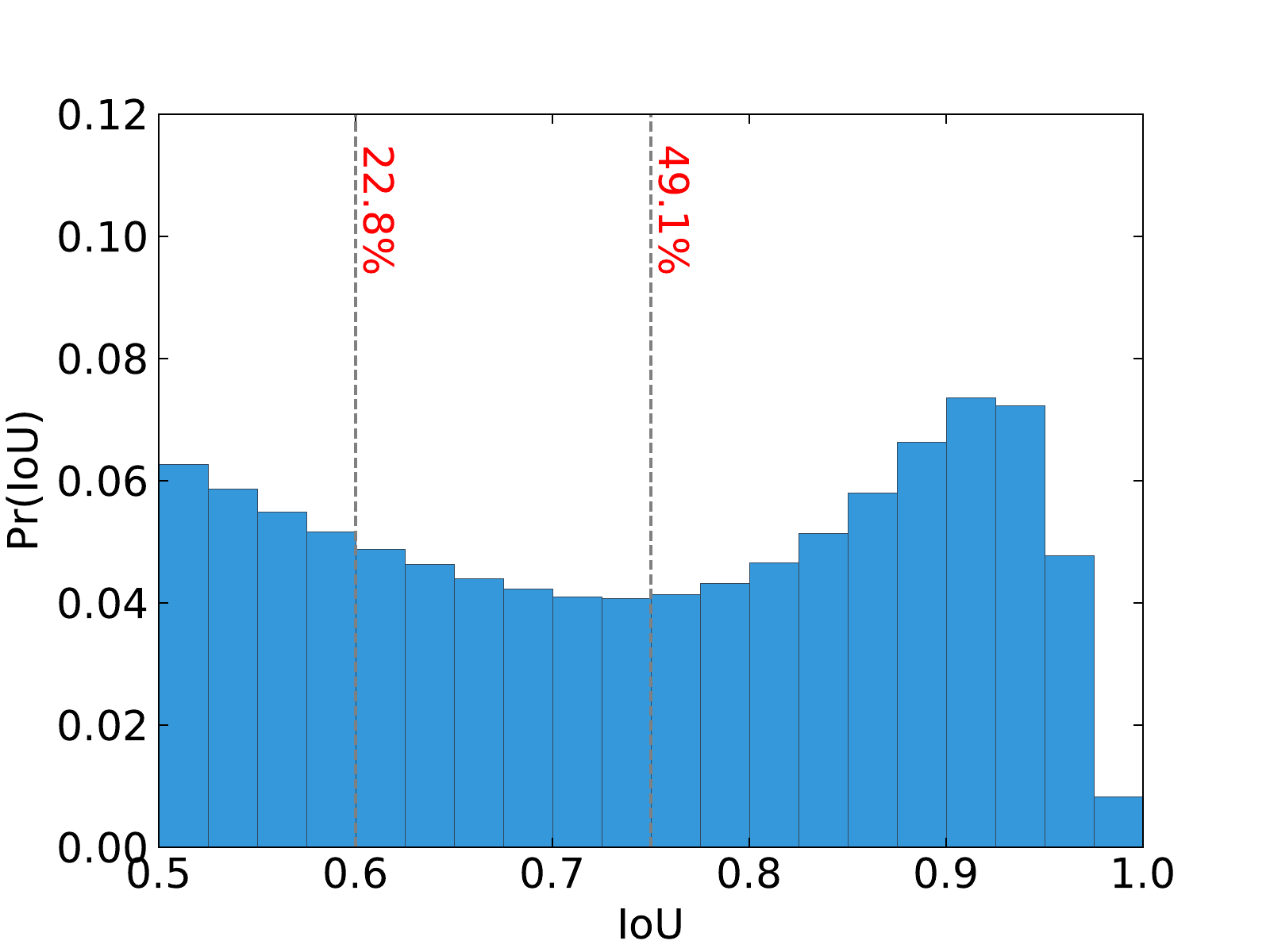}
		\caption{Training w/o sample consistency.}
		\label{fig:distribution_wo_consistency}
	\end{subfigure}
	\begin{subfigure}{0.32\textwidth}
		\centering
		\includegraphics[width=1\textwidth, height=0.7\textwidth]{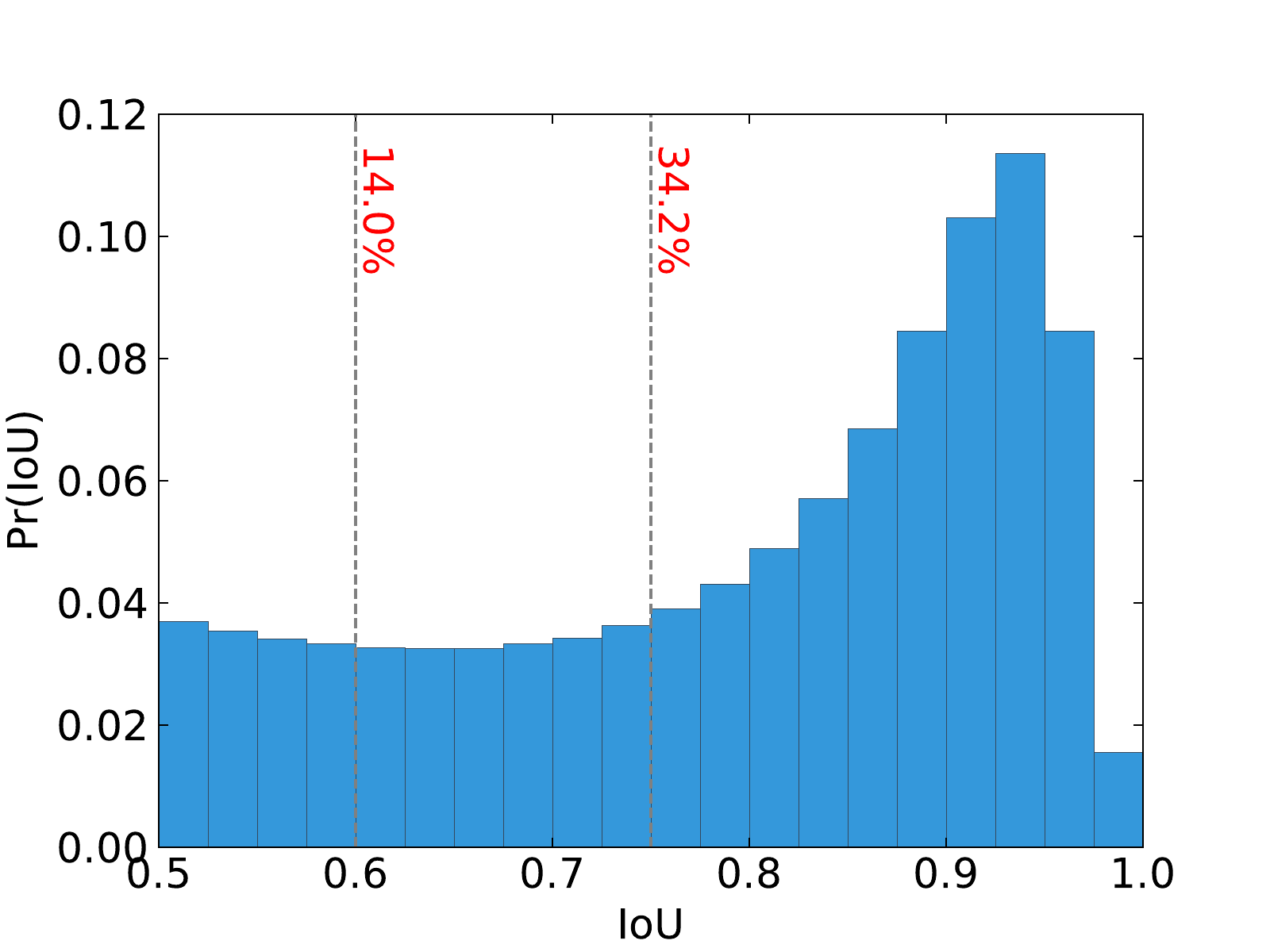}
		\caption{Training w/ sample consistency.}
		\label{fig:distribution_w_consistency}
	\end{subfigure}
	\begin{subfigure}{0.32\textwidth}
		\centering
		\includegraphics[width=1\textwidth, height=0.7\textwidth]{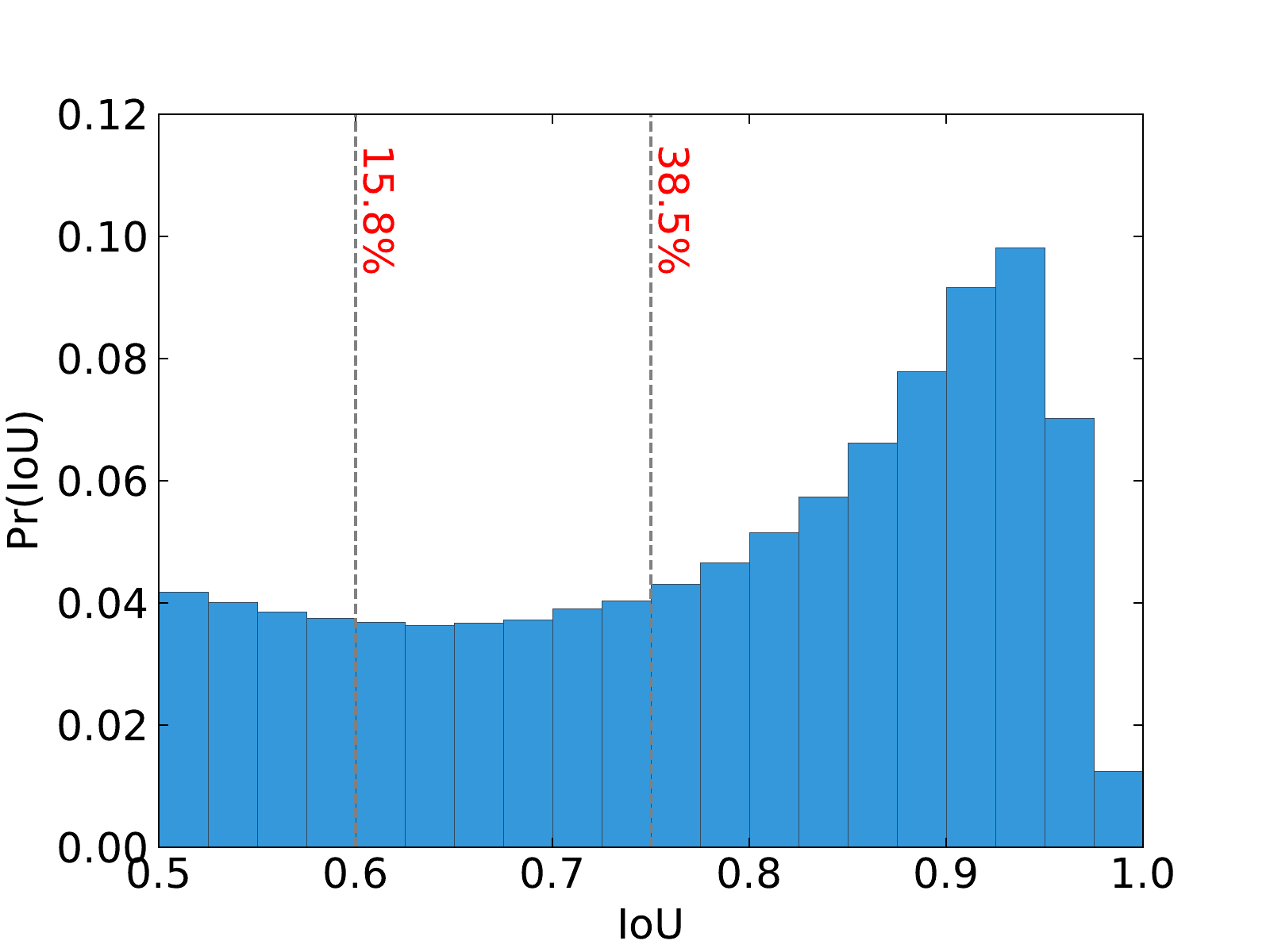}
		\caption{Inference.}
		\label{fig:distribution_inference}
	\end{subfigure}
	\caption{IoU distribution in training with/without sample consistency and inference. The red numbers show the percentages of the samples having IoU less than the corresponding threshold.}
	\label{fig:distribution}
\end{figure*}

\section{Introduction}
In recent years, instance segmentation has received considerable attention for its applications in autonomous driving \cite{neven2018towards,zhang2016instance}, robotics \cite{danielczuk2019segmenting,pathak2018learning}, surveillance \cite{Mao_2018_CVPR_Workshops,zhang2018vehicle}, and other vision tasks \cite{kim2019progressive, kim2020modality}. Given an image, instance segmentation aims to predict class labels and instance masks for objects of interest at pixel-level. Achieving an accurate and robust instance segmentation in a real-world environment is challenging: object occlusion, deformation, and scale changes are of concern.

State-of-the-art instance segmentation methods attempt to benefit from high performing object detectors, where the predicted boxes are segmented using a fully convolutional network \cite{Long_2015_CVPR}, such as Mask R-CNN \cite{He_2017_ICCV}, and PANet \cite{Liu_2018_CVPR}. For accurate object detection, Cascade R-CNN \cite{Cai_2018_CVPR} has been recently proposed, showing significant performance improvement. It consists of a sequence of detectors that progressively refine the box predictions to obtain accurate localization at the final detection stage. The capability of this detector has been extended with the addition of the mask branches for performing instance segmentation. This architecture is referred to as the Cascade Mask R-CNN \cite{cai2019cascade}, which shows significant improvement compared to non-cascade ones. Although having better performance compared to non-cascade methods, Cascade Mask R-CNN still exhibits inconsistency in training and inference sample distribution. At training time, the outputs of all the box stages are used for mask predictions; however, at inference time only the output of the last box stage is used for mask predictions. It has been shown in \cite{Cai_2018_CVPR} that the box stages produce different sample distributions since they are trained with different Intersection over Union (IoU) thresholds. Such a mismatch between training and inference sample distribution will potentially worsen the performance. 

This paper proposes an architecture referred to as Sample Consistency Network (SCNet) that ensures the IoU distribution of the samples at training time to be close to that at inference time. To this end, only the outputs of the last box stage are used for mask predictions at both training and inference. Figure \ref{fig:distribution} shows the IoU distribution of the samples going to the mask branch at training time with/without sample consistency compared to that at inference time. The COCO \cite{lin2014microsoft} \texttt{train} and \texttt{val} splits are using for training and inference, respectively. When sample consistency is not ensured, nearly half of the training samples (49.1\%) are at a low IoU region (IoU $<=$ 0.75), which are much larger than that of training with sample consistency (34.2\%) and inference (38.5\%). Overall, training with sample consistency produces closer IoU distribution between training and inference compared to that of training without sample consistency. 

Instance segmentation requires synergy among the three sub-tasks: detecting, classifying, and segmenting objects. To further reinforce the reciprocal relationships among sub-tasks, \textit{feature relay} and \textit{global context} are proposed. It is well-known that joint training of closely related tasks can improve the overall performance \cite{He_2017_ICCV,chen2019hybrid}. For instance, adding an extra mask branch to a detector improves the detection performance, although there is no direct information flow between the box and mask branches \cite{He_2017_ICCV}. It shows that the ``implicit" mutual information between detection and segmentation improves the overall performance. This paper takes this concept a step further and introduces an ``explicit" connection linking the output of the box branch to the input of the mask branch to elevate mutual information between the outputs of the two branches that ultimately enhance segmentation performance. This process is referred to as feature relay.

Common methods for detection and segmentation are performed in a region-wise manner, where the prediction is made based on features extracted from a small region by the pooling layer, such as RoIAlign \cite{He_2017_ICCV}. These layers serve as a hard attention mechanism that enables the detector to focus on the relevant region of the image. However, in a number of cases, objects are visually ambiguous when they stand alone. To overcome this limitation, SCNet relies on a global context branch to provide each object context prior for the final prediction.


\section{Related Work}
\label{sec:related_work}

\subsubsection{Instance Segmentation.} There are two main streams in instance segmentation: proposal-based and proposal-free methods. In proposal-based methods, conventional detectors \cite{girshick2015fast, vu2019cascade} generate region proposals, then instance masks are predicted within the proposed regions. DeepMask \cite{pinheiro2015learning}, SharpMask \cite{pinheiro2016learning}, and InstanceFCN \cite{dai2016instance} learn to produce segment candidates instead of bounding boxes as proposals. Li \etal extend InstanceFCN and propose FCIS for instance segmentation by introducing position-sensitive score maps \cite{Li_2017_CVPR}. In \cite{Dai_2016_CVPR}, a multi-task cascade is proposed, where the output of a sub-task is used as input of the next sub-task. He \etal present Mask R-CNN \cite{He_2017_ICCV} by appending a segmentation branch in parallel to the detection branch of Faster R-CNN \cite{NIPS2015_5638}, showing promising results. Liu \etal extend Mask R-CNN and propose PANet \cite{Liu_2018_CVPR}, which aims to enhance the feature hierarchy by adding a bottom-up path into the standard FPN \cite{Lin_2017_CVPR}. Recent advanced methods \cite{cai2019cascade,chen2019hybrid} extend the multi-stage detector Cascade R-CNN \cite{Cai_2018_CVPR} to produce high-quality instance segmentation. 

In proposal-free methods, object instances are directly identified without proposals. In \cite{zhang2016instance,Zhang_2015_ICCV}, local instance labels are predicted and integrated with a Markov Random Field (MRF) to obtain globally consistent instance labels. In \cite{arnab2016bottom}, a semantic segmentation map is first predicted, then instances are identified, relying on a Conditional Random Field (CRF) model. Bai and Urtasun propose a watershed transform network to obtain an energy map, then derive instances based on the energy levels \cite{Bai_2017_CVPR}. In \cite{Liu_2017_ICCV}, a sequence of networks is designed to predict horizontal and vertical object breakpoints, which are then merged to produce object instances. In \cite{tian2020conditional}, a dynamic instance-aware network is proposed to replace RoI opperation, leading to a compact and fast model. Xie \etal propose PolarMask, which formulates the instance segmentation problem as instance center classification and dense distance regression in a polar coordinate \cite{xie2020polarmask}. Wang \etal propose SOLO which directly predicts mask instances by assigning categories to each pixel within an instance according to the instance’s location and size \cite{Wang2020Solo}. Overall, the proposal-free methods are simple and fast; however, proposal-based methods are generally more accurate.


\begin{figure*}[!t]
	\centering
	\begin{subfigure}{0.48\textwidth}
		\centering
		\includegraphics[width=0.8\textwidth]{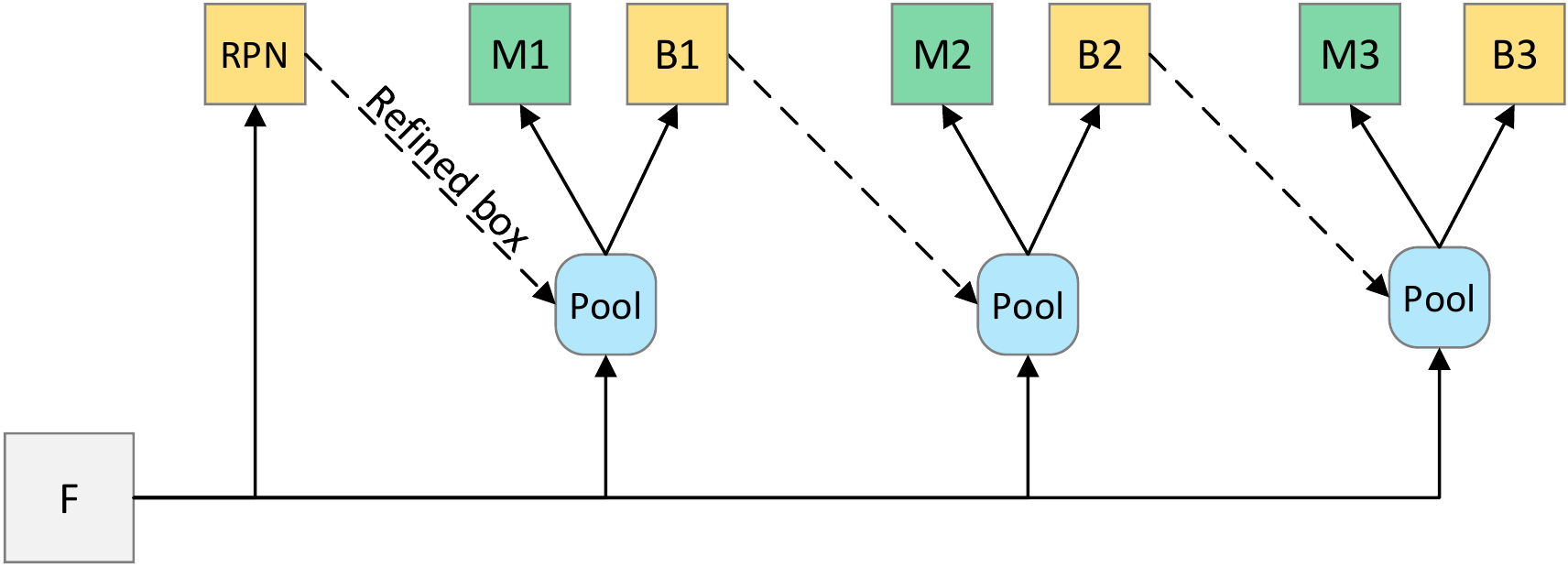}
		\caption{Cascade Mask R-CNN \cite{cai2019cascade}}
		\label{fig:compare_CMRCNN}
	\end{subfigure}
	\begin{subfigure}{0.48\textwidth}
		\centering
		\includegraphics[width=0.8\textwidth]{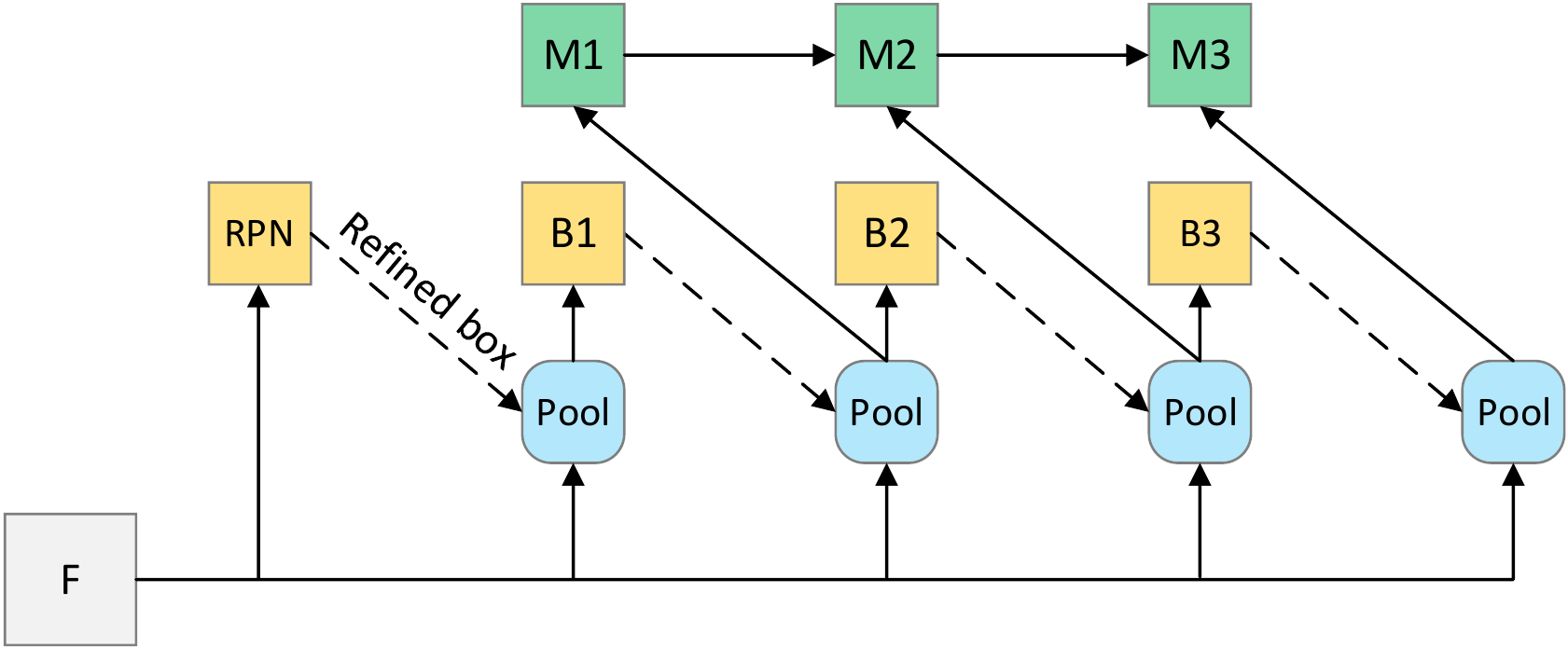}
		\caption{HTC \cite{chen2019hybrid}}
		\label{fig:compare_HTC}
	\end{subfigure}
	
	\begin{subfigure}{0.5\textwidth}
		\centering
		\includegraphics[width=0.8\textwidth]{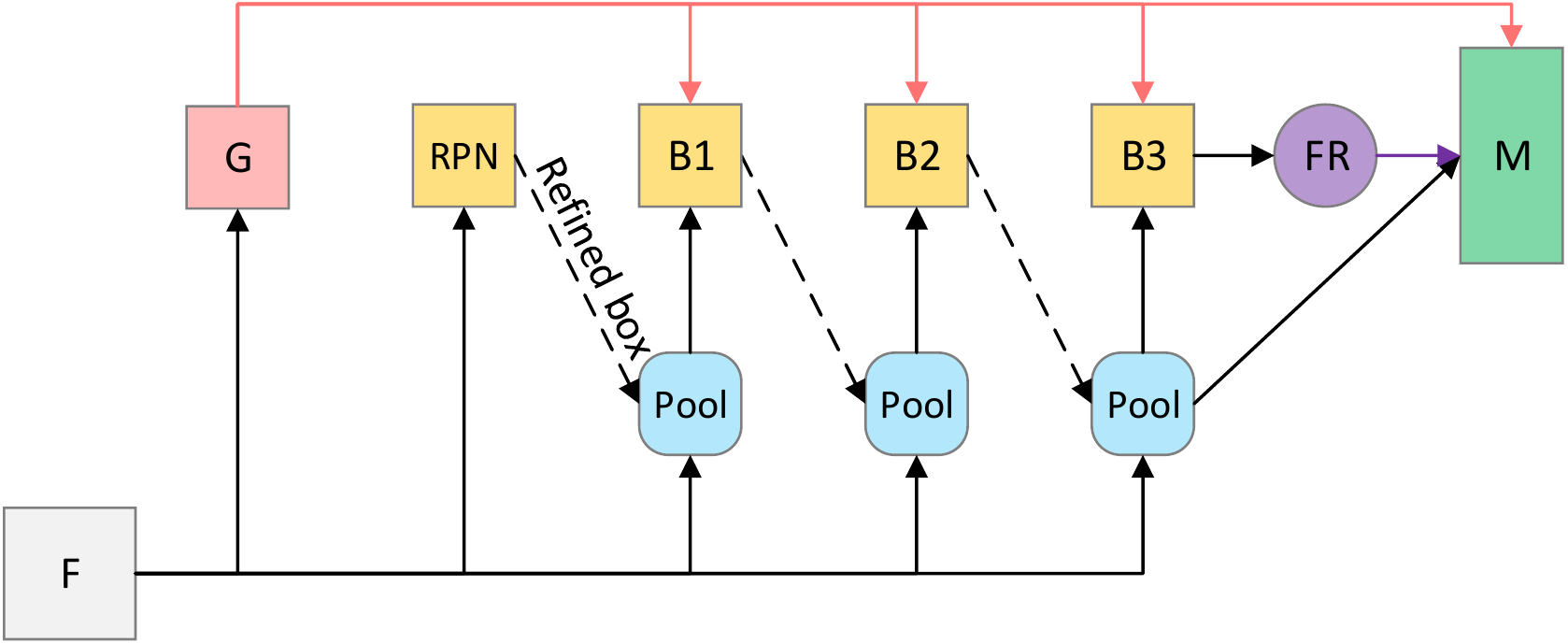}
		\caption{The proposed SCNet}
		\label{fig:compare_SCNet_5}
	\end{subfigure}
	\caption{Architecture of cascade approaches: (a) Cascade Mask R-CNN. (b) Hybrid Task Cascade (HTC). (c) the proposed SCNet. Here, ``F", ``RPN", ``Pool", ``B", ``M", ``FR" and ``G" denote image features, Region Proposal Network \cite{NIPS2015_5638}, region-wise pooling, box branch, mask branch, feature relay, and global context branch, respectively. It is noted that each box branch performs both box regression and classification. Additionally, the semantic branch \cite{chen2019hybrid}, which is not shown for a neat presentation, is applied to all cascade models for a fair comparison}
	\label{fig:compare}
\end{figure*}

\begin{figure*}[!t]
	\centering
	\begin{subfigure}{0.48\textwidth}
		\centering
		\includegraphics[width=0.8\textwidth]{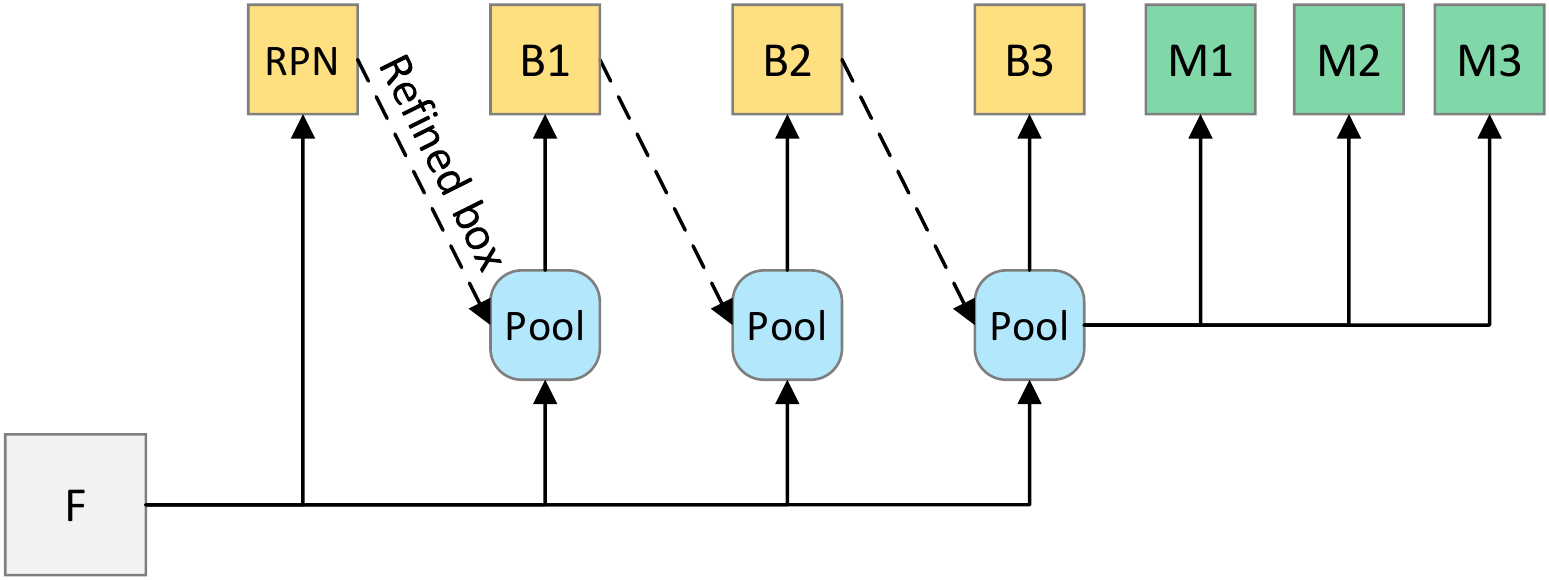}
		\caption{Sample Consistency (Naive)}
		\label{fig:compare_SCNet_1}
	\end{subfigure}
	\begin{subfigure}{0.48\textwidth}
		\centering
		\includegraphics[width=0.8\textwidth]{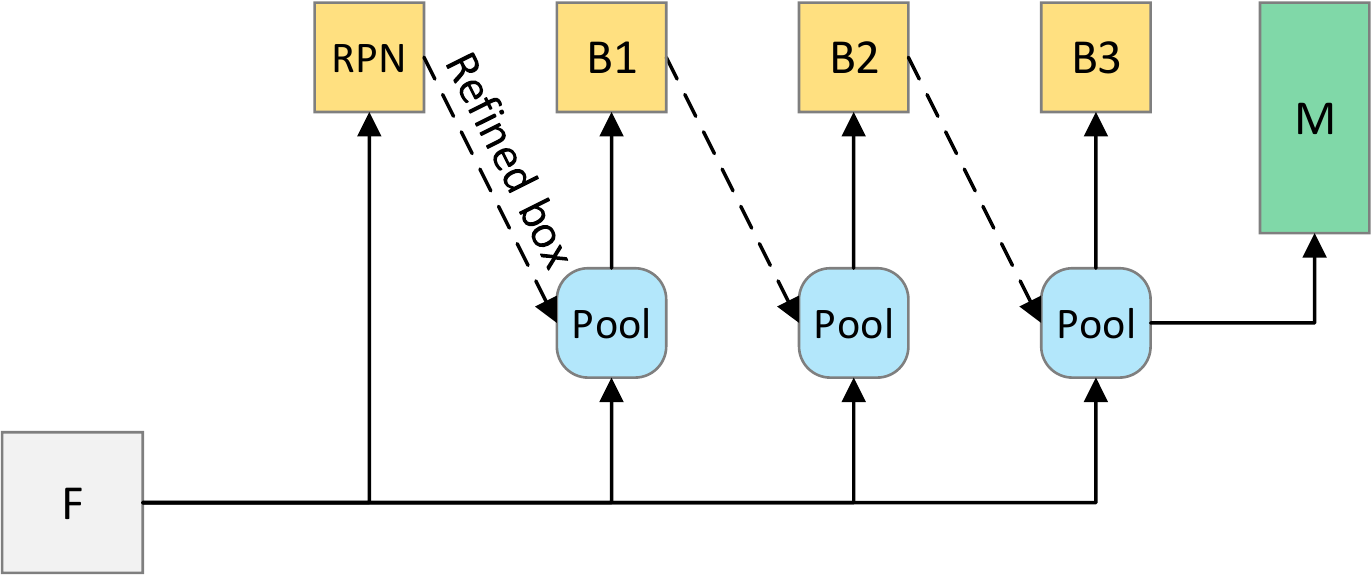}
		\caption{Sample Consistency (Effective)}
		\label{fig:compare_SCNet_2}
	\end{subfigure}
	
	\begin{subfigure}{0.48\textwidth}
		\centering
		\includegraphics[width=0.8\textwidth]{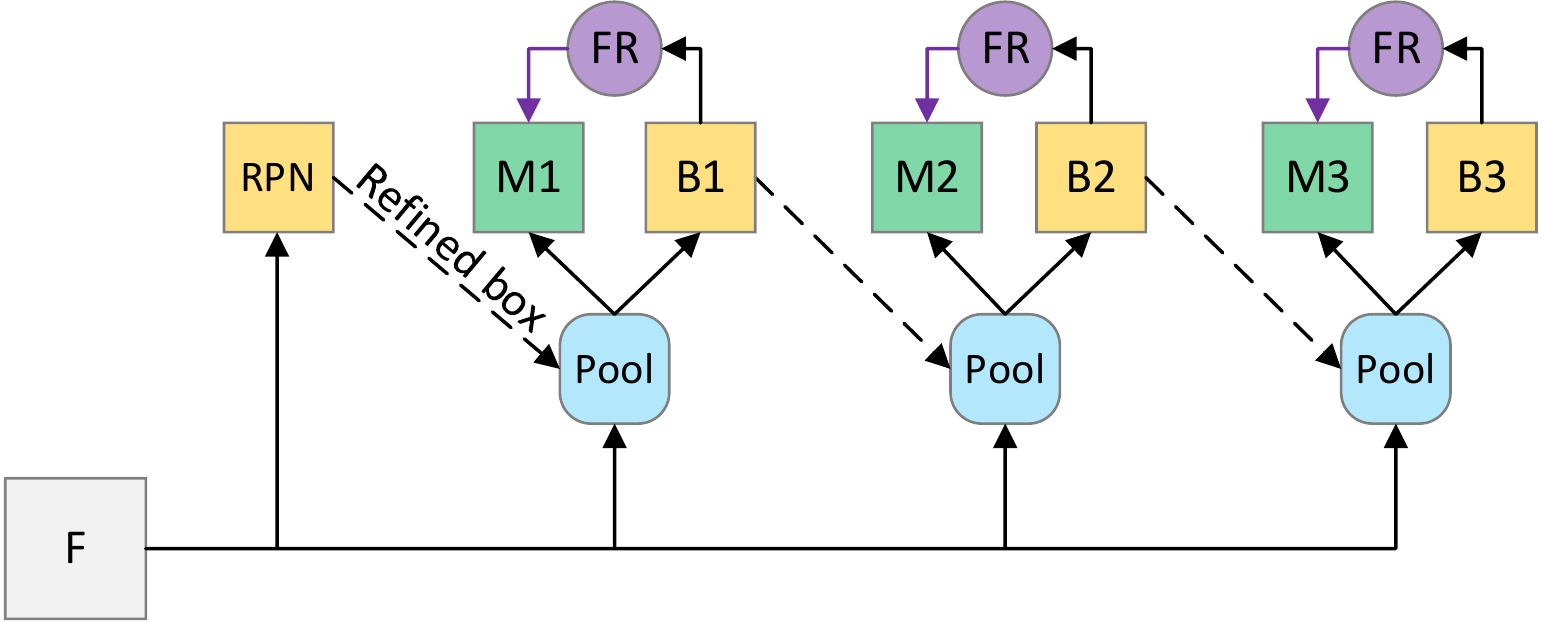}
		\caption{Feature Relay}
		\label{fig:compare_SCNet_3}
	\end{subfigure}
	\begin{subfigure}{0.48\textwidth}
		\centering
		\includegraphics[width=0.8\textwidth]{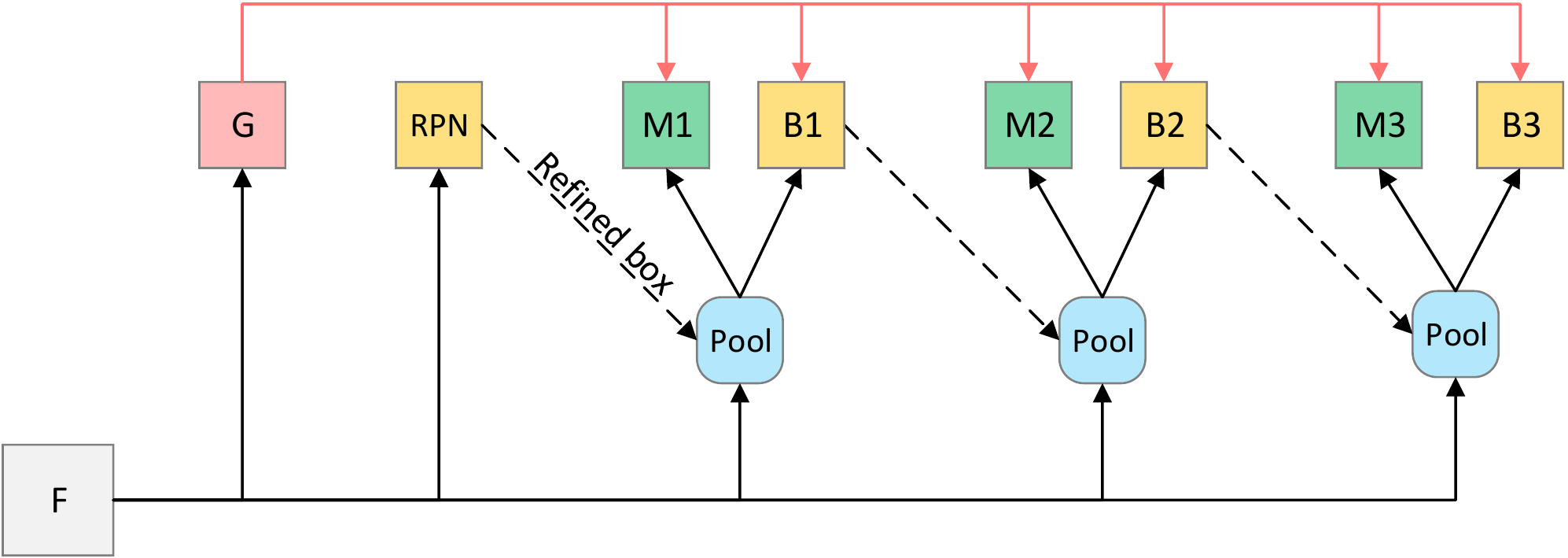}
		\caption{Global Context}
		\label{fig:compare_SCNet_4}
	\end{subfigure}
	\caption{Individual components of the proposed SCNet, which are applied to the baseline Cascade Mask R-CNN. }
	\label{fig:compare}
\end{figure*}

\subsubsection{Multi-stage Instance Segmentation.} In proposal-based approach, benefiting from the high-performing multi-stage detector, Cascade Mask R-CNN \cite{cai2019cascade} shows improvement when compared to non-cascade methods for instance segmentation. Recently, Chen et al. \cite{chen2019hybrid} have extended Cascade Mask R-CNN and propose HTC to improve segmentation by constructing mask information flow though stages and introducing a semantic branch. Although showing improvements compared to non-cascade methods, Cascade Mask R-CNN and HTC show limitation of inconsistency in training and inference sample distribution and the requirement of multi-stage mask predictions, which is not computationally efficient. The proposed SCNet differs from previous methods in that it ensures the sample consistency between training and testing time. Sample consistency is optimized to not only improve the accuracy but also improve the inference speed by avoiding the repetition of expensive operations, including mask RoI feature extraction, feature upsampling, and mask prediction. The performance of SCNet is also further improved with the incorporation of feature relay and global context information. Feature relay creates the information flow between box and mask branch. Global context provides individual objects with context prior for final prediction. The crucial difference between the proposed global context branch with previous methods in object detection  \cite{wang2018non,cao2019gcnet,qiao2020detectors} is that these methods incorporate global context in the backbone at pixel level meanwhile the proposed method incorporates global context in the detector stages (R-CNN) at instance-level. It is nontrivial to improve detection performance in R-CNN by global context since simply applying Global Convolution Network \cite{peng2017large} does not show performance gain \cite{chen2019hybrid}. It is expected that the proposed global context branch is complemented with backbone-based global context since they improve different parts of the detector.

\section{Cascade Architectures}

\subsection{Cascade Mask R-CNN}
Cascade Mask R-CNN is the combination of the high-performing detector Cascade R-CNN and the popular segmentation method Mask R-CNN. Figure \ref{fig:compare_CMRCNN} illustrates the architecture of a 3-stage Cascade Mask R-CNN. Mathematically, Cascade Mask R-CNN can be formulated as follows.
\begin{equation}
\begin{aligned}
\boldsymbol{x}_t^{box} &\leftarrow \mathcal{P}(\boldsymbol{x}, \boldsymbol{b}_{t -1 }), &&\boldsymbol{b}_t \leftarrow \mathcal{B}_t(\boldsymbol{x}_t^{box}), \\
\boldsymbol{x}_t^{mask} &\leftarrow \mathcal{P}(\boldsymbol{x}, \boldsymbol{b}_{t -1 }), &&\boldsymbol{m}_t \leftarrow \mathcal{M}_t(\boldsymbol{x}_t^{mask}).
\end{aligned}
\end{equation}
Here, $\boldsymbol{x}$ is the feature maps from a convolutional backbone network. At stage $t$, a region-wise pooling operator $\mathcal{P}$ extracts box features $\boldsymbol{x}_t^{box}$ and mask features $\boldsymbol{x}_t^{mask}$ based on the backbone features and predicted boxes (or region proposals) at the previous stage $\boldsymbol{b}_{t-1}$. The predicted boxes $\boldsymbol{b}_t$ and masks $\boldsymbol{m}_t$ are derived from the box branch $\mathcal{B}_t$ and mask branch $\mathcal{M}_t$, respectively. 

Even though performing better than other non-cascade methods, Cascade Mask R-CNN exhibits two main limitations in its architecture as follows. First, the mask predictions at training and inference come from different distributions. At training time, the outputs of all the box stages are used for mask predictions; however, at inference time only the output of the last box stage is used for mask predictions. This is because at inference time, the mask ensemble requires the mask predictions upon the same RoI locations. Although using multiple box stages for mask prediction improves the sample diversity \cite{Cai_2018_CVPR}, the proposed SCNet shows that making the training sample distribution close to that of inference further improves the performance. Second, the mask branches are isolated without direct connections, and inaccurate mask predictions are made on intermediate noisy boxes, as shown in Figure \ref{fig:compare_CMRCNN}. The effectiveness of Cascade Mask R-CNN mainly stems from the high-performing detector and the ensemble of multiple isolated mask branches.

\subsection{Hybrid Task Cascade}
To alleviate the problems of Cascade Mask R-CNN, Hybrid Task Cascade (HTC) \cite{chen2019hybrid} introduces interleaved execution between box and mask branches and direct information flow through the mask branches. The pipeline of HTC can be described as follows.
\begin{equation}
\begin{aligned}
\boldsymbol{x}_t^{box} &\leftarrow \mathcal{P}(\boldsymbol{x}, \boldsymbol{b}_{t - 1 }), \hspace{0.2cm} \boldsymbol{b}_t \leftarrow \mathcal{B}_t(\boldsymbol{x}_t^{box}), \\
\boldsymbol{x}_t^{mask} &\leftarrow \mathcal{P}(\boldsymbol{x}, \boldsymbol{b}_{t}), \hspace{0.15cm}\boldsymbol{m}_t \leftarrow \mathcal{M}_t(\mathcal{F}(\boldsymbol{x}_t^{mask}, \boldsymbol{m}_{1:t-1})).
\end{aligned}
\end{equation}

Here, HTC performs {interleaved execution} to leverage the observation that the boxes are more accurate after box regression, where the segmentation step is based on the output of detection step $\boldsymbol{b}_{t}$ instead of $\boldsymbol{b}_{t-1}$. Besides, there is a direct information flow through the mask branches. In concrete, the current backbone features $\boldsymbol{x}_t^{mask}$ are combined with the accumulated mask features from the previous stages $\boldsymbol{m}_{1:t-1}$ by a fusion operation $\mathcal{F}$. Here, $\boldsymbol{m}_{1:t-1}$ denotes the accumulated mask features taken from stage $1$ to stage $t-1$. 

To a certain extent, the interleaved execution and mask information flow alleviate the problems in Cascade Mask R-CNN; however, these ideas still have limitations to be addressed. First, the sample inconsistency remains unsolved. Second, HTC is still constrained by multi-stage mask predictions and mask ensemble. It requires multiple RoI feature extractors, upsamplers, and predictors, and practically, they are resource-consuming.

\section{The proposed SCNet}


\subsection{Sample Consistency}
The proposed SCNet introduces sample consistency that ensures the consistency in the sample distribution at training and inference. Two versions of sample consistency are considered: naive and effective sample consistency. The naive sample consistency moves all the mask branches after the last box stage and the output of the last box stage is used for extracting mask features for all mask branches at both training and inference (Figure \ref{fig:compare_SCNet_1}). Although the sample consistency is attained, it still requires the repetition of computationally expensive operations, such as RoI feature extraction, feature upsampling, and mask prediction. Computational efficiency is an important measure in segmentation, which is usually used as a front-end task for many other tasks. To speed up the network, the effective sample consistency is proposed to use a single deep mask branch instead of multiple shallow ones (Figure \ref{fig:compare_SCNet_2}), which is used by default in SCNet. In detail, the common three 4-convolution mask branches are ``stacked" to be a sequence of 12 consecutive convolution layers. Since the mask branches is deep, a skip connect is used after every two convolution layers. Effective sample consistency avoids the repetition of expensive operations since it relies on a single mask branch. Beside ensuring sample consistency and speeding up the network, the proposed method also addresses the problem of Cascade Mask R-CNN in that all the mask branches are isolated without direct connection, and the problem of mask predictions on intermediate noisy boxes. 

\subsection{Feature Relay and Global Context}
\begin{figure}[t]
	\centering
	\includegraphics[width=0.95\columnwidth]{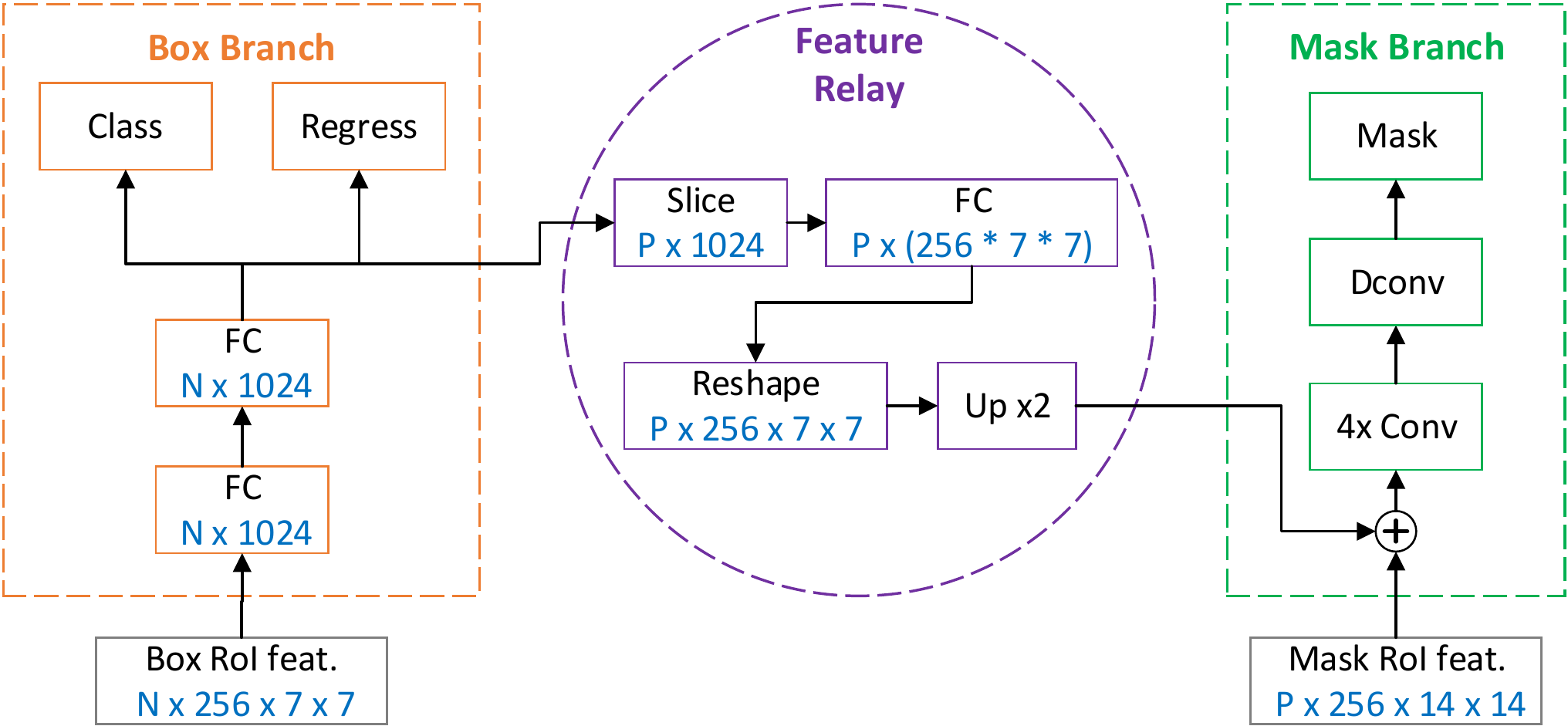}
	\caption{Architecture of feature relay. Here, ``N" and ``P" denotes the number of total samples and positive samples, respectively. ``FC", ``$\times$4 Conv", and ``Dconv" denote fully connected layer, four consecutive convolution layers, and a deconvolution layer, respectively}
	\label{fig:feature_adaptation}
\end{figure}

Feature relay and global context strengthen the relationships among classifying, detecting and segmenting sub-tasks. Motivated by the observation that ``implicit" mutual information between the box and mask branches improves the overall performance, the feature relay ``explicitly" incorporates the box features with the mask features to improve the mask prediction. Feature relay exploits the relationship between detection and segmentation sub-tasks such that the box features provide the mask branch the prior for the mask prediction, and the mask prediction supervises (refines) the box features via back-propagation. This tightly coupled relationship between detection and segmentation sub-tasks leads to the performance gain. Figure \ref{fig:feature_adaptation} shows a detailed architecture of the feature relay module. In concrete, the output features of the box branch are first sliced to obtain the ones w.r.t. positive samples then fed into a fully connected layer to align box feature space with mask feature space. The box features, which are in vector-form, are reshaped to matrix-form and upsampled before being fused with the mask features by element-wise summation. It is noted that feature relay only fuses the box and mask features at the same stage since they share the common RoI locations. When feature relay is combined with sample consistency, it is only applied to the last stage, as shown in Figure \ref{fig:compare_SCNet_5}.

\begin{figure}[t]
	\centering
	\includegraphics[width=0.95\columnwidth]{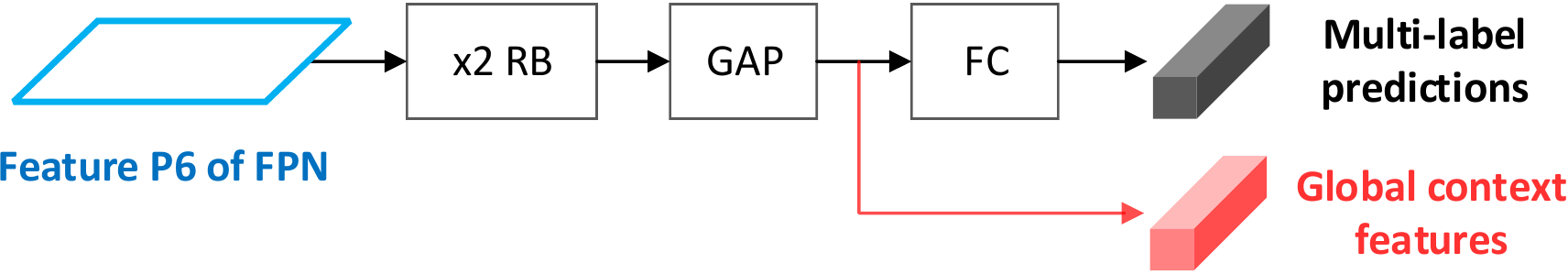}
	\caption{Global context branch takes the top-level features (P6) of FPN \cite{Lin_2017_CVPR} as the input and produces multi-label predictions and global context features. Here, ``$\times$2 RB`` and ``GAP" denotes two residual blocks and a global average pooling layer, respectively}
	\label{fig:global_context}
\end{figure}

The global context branch takes as input the backbone features and outputs the multi-label predictions and global context features. Figure \ref{fig:global_context} shows the global context branch on the FPN backbone features. First, the top-level features of FPN are fed into two consecutive residual blocks, followed by global average pooling. The pooled features are used in two sub-branches, which are multi-label classification and global context features. The multi-label predictions are supervised by all the known objects in the image, and the context features are used to fuse with box and mask features, as shown in Figure \ref{fig:compare_SCNet_4}.

\subsection{Training}


The proposed SCNet can be trained in an end-to-end manner using multi-task loss as follows:
\begin{equation}
\mathcal{L} = \sum_{t=1}^{T}\alpha_t(\mathcal{L}_t^{cls} + \mathcal{L}_t^{reg}) + \beta\mathcal{L}^{mask} + \gamma\mathcal{L}^{sema} + \lambda\mathcal{L}^{glbctx}.
\end{equation}
Here, $\mathcal{L}^{cls}$, $\mathcal{L}^{reg}$, $\mathcal{L}^{mask}$, and $\mathcal{L}^{sema}$ are the losses of classification, regression, mask prediction, and semantic prediction, respectively. The concrete loss types and loss weights (\ie $\alpha_t$ and $\gamma$) are referred to in \cite{chen2019hybrid} without any modifications. Since effective sample consistency uses only one mask branch, the mask loss weight is re-weighted to equal the summation of stage-wise loss weights: 
\begin{equation}
\beta = \sum_{t=1}^{T}\alpha_t.
\label{eq:loss_balance}
\end{equation}
Besides, SCNet introduce a new global context loss $\mathcal{L}^{glbctx}$, which performs multi-label classification and is implemented using binary cross entropy. 

\section{Experiments}
\label{sec:experiments}

\subsection{Implementation Details}
The default model consists of 3 cascading stages with the ResNet FPN \cite{Lin_2017_CVPR} being the backbone network. The stage loss weights and semantic loss weight, which are adopted from \cite{chen2019hybrid}, are set to $\alpha = [1, 0.5, 0.25]$ and $\gamma=0.2$, respectively. The global context loss weight is set to $\lambda = 3$. 

In all experiments, the long edge and short edge of the images are resized to 1333 and 800, respectively, without changing the aspect ratio. No data augmentation is used except for standard horizontal image flipping. PyTorch \cite{paszke2017automatic} and MMDetection \cite{mmdetection} are used for implementation. The models are trained with 8 GPUs with a batch size of 16 (two images per GPU) for 20 epochs using SGD optimizer. The learning rate is initialized to 0.02 and divided by 10 after 16 and 19 epochs, respectively. It takes about one day for the models to converge on 8 Tesla V100 GPUs. 

During test time, object proposals are progressively refined by box branches of different stages. The final classification score for each detected box is obtained by averaging the scores of multiple classifiers, referred to in Cascade R-CNN \cite{Cai_2018_CVPR}. Only the detected boxes with classification scores higher than a threshold of 0.001 are segmented by the mask branch. 

Detection and segmentation results are evaluated with the standard COCO-style Average Precision (AP) metric. The runtime is measured on a single Tesla V100 GPU.

\begin{table*}[]
	\caption{Benchmarking results between the proposed SCNet and other state-of-the-art methods on COCO \texttt{test-dev}. Here, AP and AP$^\text{bb}$ are the mask AP and box AP, respectively. The semantic branch \cite{chen2019hybrid} is used for all cascade models.}
	\centering
	\begin{tabular}{@{}clccccccccc@{}}
		\toprule
		Type                                                                               & \quad Method             & Backbone                     & AP & AP$_{50}$ & AP$_{75}$ & AP$_{S}$  & AP$_{M}$  & AP$_{L}$ & AP$^\text{bb}$ & Speed (fps) \\ \midrule
		\multirow{6}{*}{\begin{tabular}[c]{@{}c@{}}None-\\ cascade\\ methods\end{tabular}} & \quad Mask R-CNN         & \multirow{5}{*}{ResNet-50}   & 35.6          & 57.6          & 38.0          & 18.9          & 38.4          & 46.4          & 38.9          & 11.3           \\
		& \quad PANet              &                              & 36.6          & 58.0          & 39.9          & 16.3          & 38.1          & 52.4          & 41.2          & -              \\
		& \quad LevelSet R-CNN     &                              & 36.4          & -             & -             & -             & -             & -             & -             & -              \\
		& \quad BlendMask          &                              & 37.0          & 58.9          & 39.7          & 17.3          & 39.4          & 52.5          & -             & -              \\
		& \quad BMask R-CNN        &                              & 35.9          & 57.0          & 38.6          & 15.8          & 37.6          & 52.2          & -             & -              \\ \cmidrule(l){2-11} 
		& \quad D2Det              & ResNet-101                   & 40.2          & 61.5          & 43.7          & -             & -             & -             & 45.4          & -              \\ \midrule
		\multirow{9}{*}{\begin{tabular}[c]{@{}c@{}}Cascade\\ methods\end{tabular}}         & \quad Cascade Mask R-CNN & \multirow{3}{*}{ResNet-50}   & 37.9          & 59.8          & 40.8          & 20.2          & 40.2          & 50.2          & 43.7          & 4.5            \\
		& \quad HTC                &                              & 38.5          & 60.1          & 41.7          & 20.4          & 40.6          & 51.2          & 43.6          & 4.5            \\
		& \quad SCNet (ours)       &                              & \textbf{40.2} & \textbf{62.3} & \textbf{43.4} & \textbf{22.4} & \textbf{42.8} & \textbf{53.4} & \textbf{45.0} & \textbf{6.2}   \\ \cmidrule(l){2-11} 
		& \quad Cascade Mask R-CNN & \multirow{3}{*}{ResNet-101}  & 39.2          & 61.3          & 42.4          & 20.9          & 41.7          & 52.2          & 45.3          & 4.4            \\
		& \quad HTC                &                              & 39.7          & 61.8          & 43.0          & 20.9          & 42.4          & 53.0          & 45.1          & 4.4            \\
		& \quad SCNet (ours)       &                              & \textbf{41.3} & \textbf{63.9} & \textbf{44.8} & \textbf{22.7} & \textbf{44.1} & \textbf{55.2} & \textbf{46.4} & \textbf{5.8}   \\ \cmidrule(l){2-11} 
		& \quad Cascade Mask R-CNN & \multirow{3}{*}{ResNeXt-101} & 40.9          & 63.7          & 44.2          & 22.4          & 43.5          & 54.2          & 47.3          & 3.7            \\
		& \quad HTC                &                              & 41.3          & 63.9          & 44.8          & 22.7          & 44.0          & 54.7          & 47.2          & 3.7            \\
		& \quad SCNet (ours)       &                              & \textbf{42.7} & \textbf{65.7} & \textbf{46.4} & \textbf{24.1} & \textbf{45.7} & \textbf{56.3} & \textbf{48.3} & \textbf{4.6}   \\ \bottomrule
	\end{tabular}
	\label{tab:benchmark}
\end{table*}

\begin{table*}[t]
	\caption{Ablation study of the proposed SCNet on COCO \texttt{val}, the baseline is Cascade Mask R-CNN \cite{cai2019cascade} with semantic \cite{chen2019hybrid}}
	\centering
	\setlength{\tabcolsep}{3.5pt}
	\begin{tabular}{ccccccccc}
		\toprule
		Cascade Mask R-CNN  & Sample Consistency & Feature Relay & Global Context    & AP   & AP$_{50}$ & AP$_{75}$ & AP$^\text{bb}$ & Speed (fps) \\ \midrule
		\checkmark &                 &              &            & 37.4 & 59.0 & 40.0 & 43.3   & 4.5   \\
		\checkmark & \checkmark      &              &            & 38.8 & 59.8 & 41.7 & 43.5   & \textbf{6.5}   \\
		\checkmark &                 & \checkmark   &            & 38.0 & 59.2 & 40.8 & 43.4   & 4.1   \\
		\checkmark &                 &              & \checkmark & 38.3 & 59.8 & 41.2 & 44.5   & 4.4     \\
		\checkmark & \checkmark      & \checkmark   &            & 39.0 & 60.0 & 41.9 & 43.7   & 6.3   \\
		\checkmark & \checkmark      & \checkmark   & \checkmark & \textbf{39.8} & \textbf{61.4} & \textbf{42.7} & \textbf{44.6}   & 6.2   \\ \midrule
		\multicolumn{4}{c}{Overall Improvement}                  & \textbf{+2.4}  & \textbf{+2.4}  & \textbf{+2.7}  & \textbf{+1.3}    & \textbf{+1.7}   \\ \bottomrule
	\end{tabular}
	\label{tab:component-wise}
\end{table*}

\subsection{Benchmarking Results}
The performance of SCNet is compared to that of recent state-of-the-art instance segmentation methods, including Cascade Mask R-CNN \cite{cai2019cascade} and Hybrid Task Cascade (HTC) \cite{chen2019hybrid}. For a fair comparison, the semantic branch, referred to in \cite{chen2019hybrid}, is adopted for all the cascade models. Besides, SCNet is also benchmarked with other non-cascade models, including Mask R-CNN \cite{He_2017_ICCV} and PANet \cite{Liu_2018_CVPR}, LevelSet R-CNN \cite{homayounfar2020levelset}, BlendMask \cite{chen2020blendmask}, BMask R-CNN \cite{cheng2020boundary}, and D2Det \cite{cao2020d2det}. 

\begin{figure*}[t]
	\centering
	\begin{tabular}{c@{ }c}
		\rotatebox[origin=c]{90}{CM R-CNN} &
		\begin{subfigure}{0.14\textwidth}
			\centering
			\includegraphics[width=1\textwidth, height=0.12\textheight]{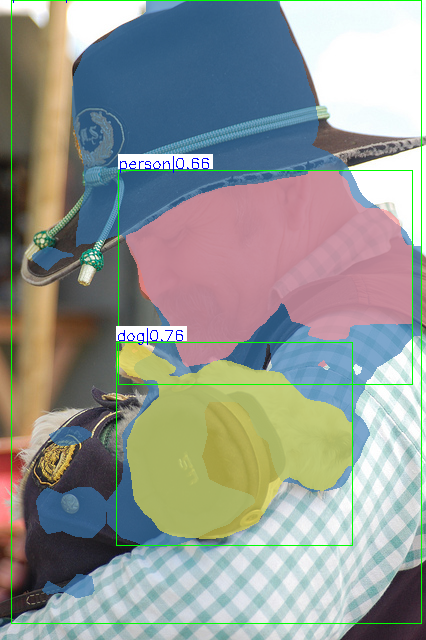}
		\end{subfigure}
		\begin{subfigure}{0.16\textwidth}
			\centering
			\includegraphics[width=1\textwidth, height=0.12\textheight]{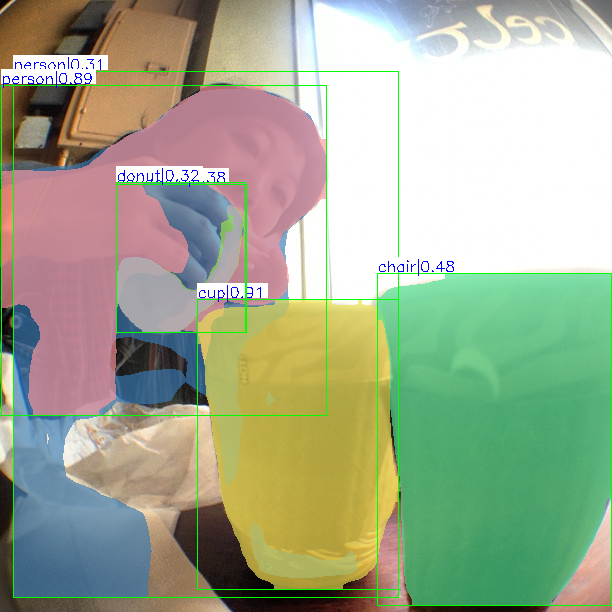}
		\end{subfigure}
		\begin{subfigure}{0.215\textwidth}
			\centering
			\includegraphics[width=1\textwidth, height=0.12\textheight]{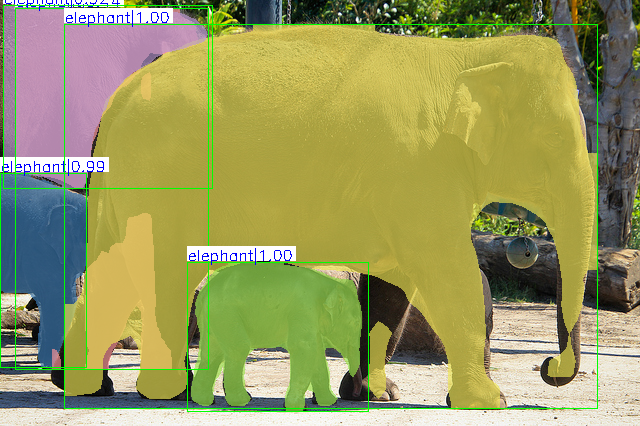}
		\end{subfigure}
		\begin{subfigure}{0.215\textwidth}
			\centering
			\includegraphics[width=1\textwidth, height=0.12\textheight]{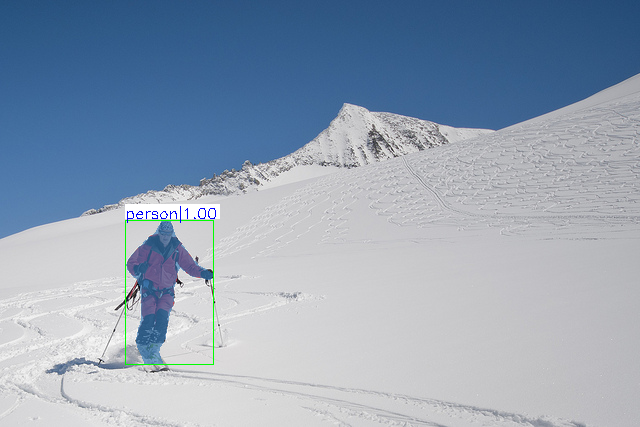}
		\end{subfigure}
		\begin{subfigure}{0.215\textwidth}
			\centering
			\includegraphics[width=1\textwidth, height=0.12\textheight]{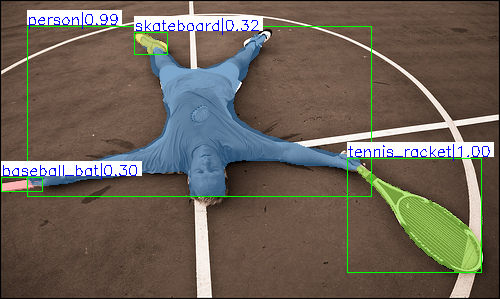}
		\end{subfigure}
	\end{tabular}
	
	\centering
	\begin{tabular}{c@{ }c}
		\rotatebox[origin=c]{90}{HTC} &
		\begin{subfigure}{0.14\textwidth}
			\centering
			\includegraphics[width=1\textwidth, height=0.12\textheight]{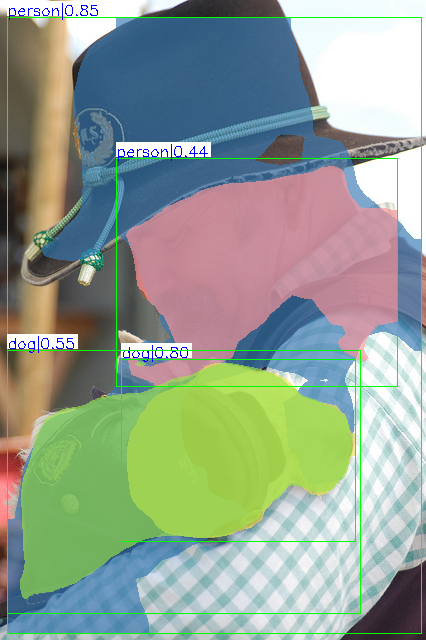}
		\end{subfigure}
		\begin{subfigure}{0.16\textwidth}
			\centering
			\includegraphics[width=1\textwidth, height=0.12\textheight]{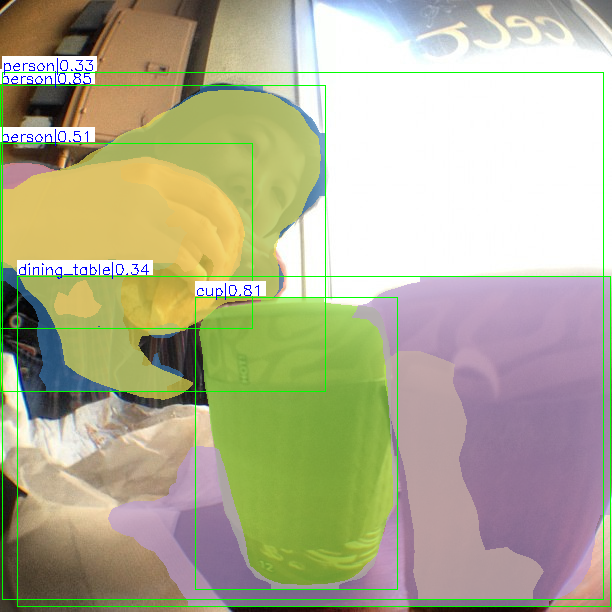}
		\end{subfigure}
		\begin{subfigure}{0.215\textwidth}
			\centering
			\includegraphics[width=1\textwidth, height=0.12\textheight]{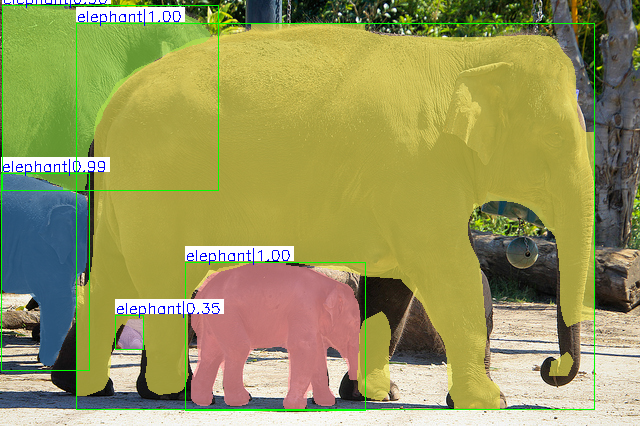}
		\end{subfigure}
		\begin{subfigure}{0.215\textwidth}
			\centering
			\includegraphics[width=1\textwidth, height=0.12\textheight]{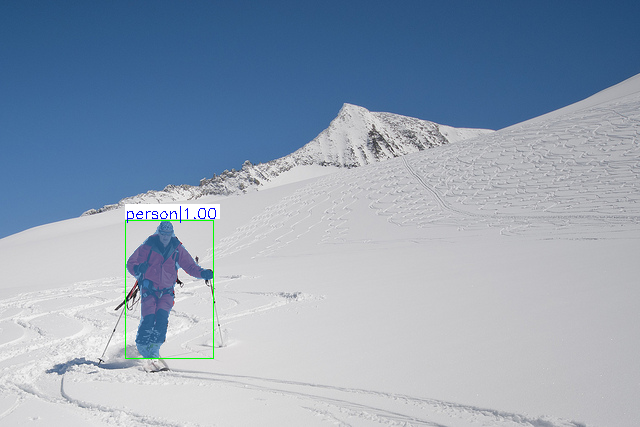}
		\end{subfigure}
		\begin{subfigure}{0.215\textwidth}
			\centering
			\includegraphics[width=1\textwidth, height=0.12\textheight]{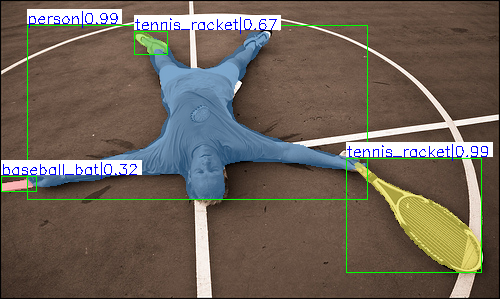}
		\end{subfigure}
	\end{tabular}
	
	\centering
	\begin{tabular}{c@{ }c}
		\rotatebox[origin=c]{90}{SCNet} &
		\begin{subfigure}{0.14\textwidth}
			\centering
			\includegraphics[width=1\textwidth, height=0.12\textheight]{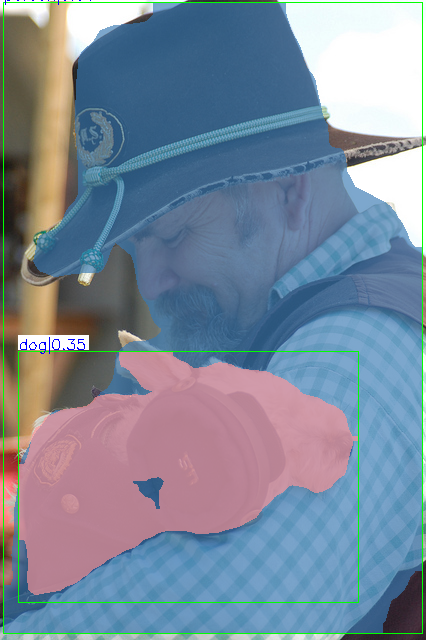}
		\end{subfigure}
		\begin{subfigure}{0.16\textwidth}
			\centering
			\includegraphics[width=1\textwidth, height=0.12\textheight]{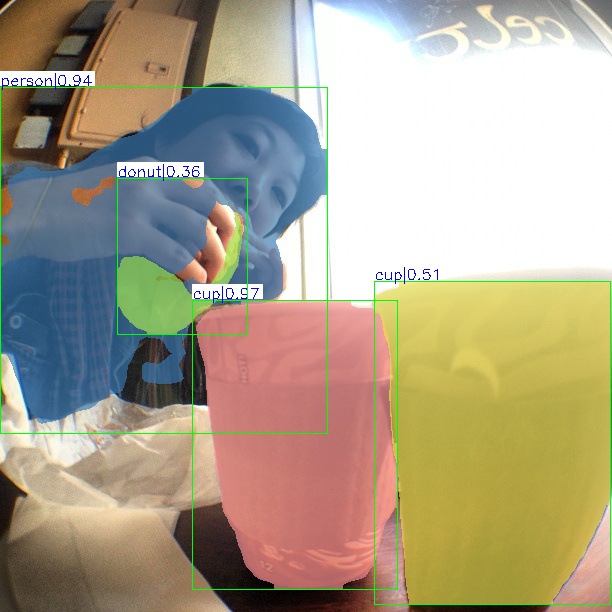}
		\end{subfigure}
		\begin{subfigure}{0.215\textwidth}
			\centering
			\includegraphics[width=1\textwidth, height=0.12\textheight]{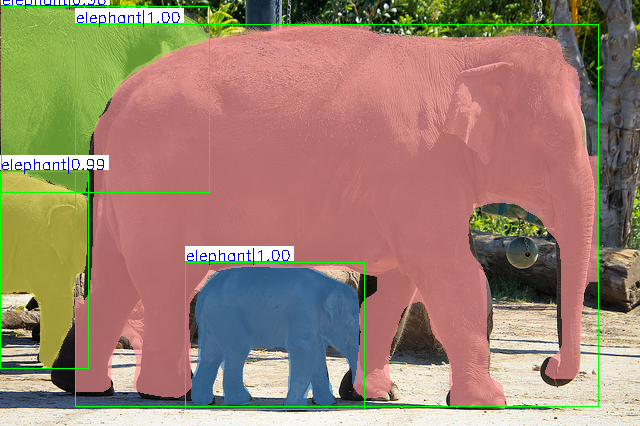}
		\end{subfigure}
		\begin{subfigure}{0.215\textwidth}
			\centering
			\includegraphics[width=1\textwidth, height=0.12\textheight]{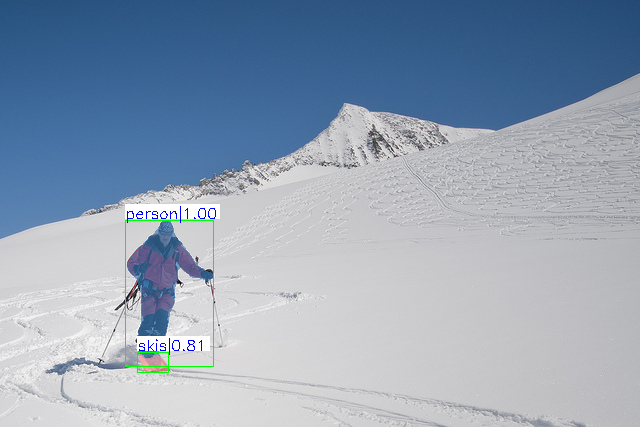}
		\end{subfigure}
		\begin{subfigure}{0.215\textwidth}
			\centering
			\includegraphics[width=1\textwidth, height=0.12\textheight]{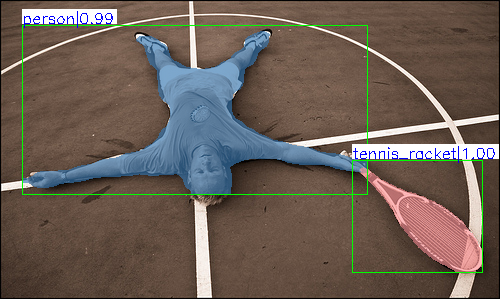}
		\end{subfigure}
	\end{tabular}
	\caption{Qualitative comparison between the proposed SCNet and other methods on COCO \texttt{val} (\textbf{zoom-in for best view}). The proposed SCNet produces more accurate the number of instances with better segmented masks.}
	\label{fig:qualitative}
\end{figure*}

Table \ref{tab:benchmark} reports the benchmarking results of the state-of-the-art segmentation methods. Overall, the cascade models show better box AP and mask AP than those of  the non-cascade ones. Among cascade models, the proposed SCNet achieves the best performance in not only box AP and mask AP but also inference speed, irrespective of the backbone strength. In particular, with the default setting of backbone ResNet-50, the proposed SCNet achieves 1.3 and 2.3 points box AP and mask AP improvements, respectively. The mask AP at different IoU thresholds (AP$_{50}$, AP$_{75}$) and object scales (AP$_S$, AP$_M$, AP$_L$) are also consistently higher than other methods. Regarding inference speed, SCNet runs at 6.2 fps, which is 1.7 fps (38\%) faster than Cascade Mask R-CNN and HTC. When applying better backbones of ResNet-101 or ResNeXt-101, SCNet also outperforms other methods among the benchmarking metrics, demonstrating the effectiveness of the proposed method. Qualitatively, Figure \ref{fig:qualitative} shows the visual comparison of the proposed SCNet with other cascade models. It is clear that SCNet produces more accurate the number of instances with better-segmented masks.

\subsection{Ablation Study}
\subsubsection{Component-wise Analysis.} To demonstrate the effectiveness of the proposed SCNet, a comprehensive component-wise analysis is performed, where different components are omitted. The results are reported in Table \ref{tab:component-wise}. Here, the baseline is Cascade Mask R-CNN \cite{cai2019cascade} with the semantic branch \cite{chen2019hybrid} being applied, yielding the box AP and mask AP of 43.3 and 37.4, respectively. When sample consistency is ensured, the mask AP increases significantly to 38.8. The inference speed is also improved to 6.5 fps. When feature relay and global context are applied, the mask AP improvements are 0.6 and 0.9 points, respectively. When sample consistency and feature relay is combined, the mask AP increase from 38.8 to 39.0. The improvement of 0.2 points is because of the number of feature relay modules reduces from 3 to 1 when sample consistency is combined with feature relay. The combination of all components brings the best mask and box AP of 39.8 and 33.6, respectively. Overall, SCNet achieves respectively 2.4 points, 1.3 points and 1.7 fps improvements in terms of mask AP, box AP, and inference speed in comparison with the baseline.


\begin{table}[]
	\centering
	\small
	\caption{Comparison between naive and effective sample consistency (denoted as SC). Cascade Mask R-CNN is denoted as CM R-CNN.}
	\begin{tabular}{cccccc} \toprule
		CM R-CNN   & Naive SC   & Effective SC & AP   & AP bb & Speed (fps) \\ \midrule
		\checkmark &            &              & 37.4 & 43.3  & 4.5         \\
		\checkmark & \checkmark &              & 38.0 & 43.5  & 4.5         \\
		\checkmark &            & \checkmark   & 38.8 & 43.5  & 6.5         \\ \bottomrule
	\end{tabular}
	\label{tab:sample_consistency}
\end{table}

\subsubsection{Sample Consistency.} To demonstrate the effectiveness of sample consistency, the experiments of the naive and effective sample consistency are reported, as shown in Table \ref{tab:sample_consistency}. When naive sample consistency is applied, the mask AP is improved from 37.4 to 38.0 and the speed is kept unchanged of 4.5 fps. This is because the naive sample consistency still requires multiple mask branches. When effective sample consistency is used, both mask AP and inference speed are significantly improved. The inference speed is faster since it avoid the repetition of expensive operations through stages. The better mask AP shows that using a single deep network is more beneficial compared to using multiple shallow ones.


\subsubsection{Feature Relay.} The feature relay fuses the adapted box features with the mask features to achieves better mask prediction. feature relay can be seamlessly applied to common segmentation methods, such as Mask R-CNN and Cascade R-CNN. Table \ref{tab:feature_cascade} shows that feature relay can improve mask AP by 0.6 points in both Mask R-CNN and Cascade Mask R-CNN with marginal computational overhead. The box AP is comparable to the baseline since the feature relay aims to improve the mask predictions only.


\begin{table}[t]
	\small
	\centering
	\caption{Ablation study on Feature Relay.}
	\begin{tabular}{ccccc} \toprule
		Model                     & Feature Relay & AP   & AP$^\text{bb}$ & Speed (fps) \\ \midrule
		\multirow{2}{*}{Mask R-CNN}  & N                & 35.1 & \textbf{38.4}  & \textbf{7.7}   \\
		& Y      & \textbf{35.7}  & 38.3  & 7.4   \\  \midrule
		\multirow{2}{*}{Cascade Mask R-CNN} & N                & 37.4 & 43.3  & \textbf{4.5}   \\
		& Y      & \textbf{38.0} & \textbf{43.4}  & 4.1  \\ \bottomrule  
	\end{tabular}
	\label{tab:feature_cascade}
\end{table}

\begin{table}[t]
	\caption{Ablation study of the global context loss weight $\lambda$.}
	\centering
	\setlength{\tabcolsep}{5pt}
	\begin{tabular}{cccccccc} \toprule
		$\lambda$ & AP & AP$_{50}$ & AP$_{75}$ & AP$_{S}$  & AP$_{M}$  & AP$_{L}$ & AP$^\text{bb}$        \\ \midrule
		None    & 39.0          & 60.0          & 42.0          & 20.2          & 41.8        & 54.5          & 43.7          \\
		0       & 39.6          & 61.3          & 42.6          & 21.9          & 42.7        & 54.8          & 44.4          \\
		1       & 39.6          & 61.3          & 42.6          & 21.9          & 42.8        & 55.0          & 44.5          \\
		2       & 39.7          & 61.2          & \textbf{42.8} & 21.7          & 42.9        & \textbf{55.1} & \textbf{44.6} \\
		3       & \textbf{39.8} & \textbf{61.4} & 42.7          & \textbf{22.1} &\textbf{43.0}& 54.5          & \textbf{44.6} \\
		5       & 39.6          & 61.3          & 42.6          & 21.8          & 42.7        & 55.0          & 44.5          \\ \bottomrule     
	\end{tabular}
	\label{tab:global_context_weight}
\end{table}

\subsubsection{Global Context.} The effectiveness of global context branch under different settings is studied in Table \ref{tab:global_context_weight}. Here, when the global context is not used ($\lambda$ is ``None"), SCNet achieves the mask AP of 39.0. When the global context branch is used but the loss weight is set to 0, the mask AP increase to 39.6. The performance increases when the loss weight is greater than 0 and is not sensitive to the loss weight. Setting the weight to 3 achieves the best overall performance, which is 0.8 points mask AP improvement.

\section{Conclusion}
\label{sec:conclusion}
This paper introduces SCNet, a simple yet effective architecture for instance segmentation. The proposed SCNet ensures sample consistency of IoU distribution in training and inference while speeding up the network. Furthermore, SCNet strengthens the relationships of the sub-tasks by feature relay and global context. Extensive experiments on the standard COCO dataset show the effectiveness of the proposed method in multiple evaluation metrics. In concrete, while running at a faster inference speed, the proposed SCNet improves the Average Precision of the box and mask predictions by respectively 1.3 and 2.3 points compared to the strong Cascade Mask R-CNN baseline. 

\section*{Acknowledgement}
This work was partly supported by Institute for Information \& communications Technology Planning \& Evaluation (IITP) grant funded by the Korea government (MSIP) (No. 2018-0-00198), Object information extraction and real-to-virtual mapping based AR technology) and  partly supported by Institute for Information \& communications Technology Planning \& Evaluation(IITP) grant funded by the Korea government(MSIT) (2017-0-01780, The technology development for event recognition/relational reasoning and learning knowledge based system for video understanding).

\bibliography{references}

\begin{thebibliography}{41}
\providecommand{\natexlab}[1]{#1}
\providecommand{\url}[1]{\texttt{#1}}
\providecommand{\urlprefix}{URL }
\expandafter\ifx\csname urlstyle\endcsname\relax
  \providecommand{\doi}[1]{doi:\discretionary{}{}{}#1}\else
  \providecommand{\doi}{doi:\discretionary{}{}{}\begingroup
  \urlstyle{rm}\Url}\fi

\bibitem[{Arnab and Torr(2016)}]{arnab2016bottom}
Arnab, A.; and Torr, P.~H. 2016.
\newblock Bottom-up Instance Segmentation using Deep Higher-order Crfs.
\newblock \emph{arXiv:1609.02583} .

\bibitem[{Bai and Urtasun(2017)}]{Bai_2017_CVPR}
Bai, M.; and Urtasun, R. 2017.
\newblock Deep Watershed Transform for Instance Segmentation.
\newblock In \emph{CVPR}.

\bibitem[{Cai and Vasconcelos(2018)}]{Cai_2018_CVPR}
Cai, Z.; and Vasconcelos, N. 2018.
\newblock Cascade R-CNN: Delving Into High Quality Object Detection.
\newblock In \emph{CVPR}.

\bibitem[{Cai and Vasconcelos(2019)}]{cai2019cascade}
Cai, Z.; and Vasconcelos, N. 2019.
\newblock Cascade R-CNN: High Quality Object Detection and Instance
  Segmentation.
\newblock \emph{arXiv:1906.09756} .

\bibitem[{Cao et~al.(2020)Cao, Cholakkal, Anwer, Khan, Pang, and
  Shao}]{cao2020d2det}
Cao, J.; Cholakkal, H.; Anwer, R.~M.; Khan, F.~S.; Pang, Y.; and Shao, L. 2020.
\newblock D2Det: Towards High Quality Object Detection and Instance
  Segmentation.
\newblock In \emph{CVPR}.

\bibitem[{Cao et~al.(2019)Cao, Xu, Lin, Wei, and Hu}]{cao2019gcnet}
Cao, Y.; Xu, J.; Lin, S.; Wei, F.; and Hu, H. 2019.
\newblock Gcnet: Non-local networks meet squeeze-excitation networks and
  beyond.
\newblock In \emph{ICCV Workshops}.

\bibitem[{Chen et~al.(2020)Chen, Sun, Tian, Shen, Huang, and
  Yan}]{chen2020blendmask}
Chen, H.; Sun, K.; Tian, Z.; Shen, C.; Huang, Y.; and Yan, Y. 2020.
\newblock BlendMask: Top-down meets bottom-up for instance segmentation.
\newblock In \emph{CVPR}.

\bibitem[{Chen et~al.(2019{\natexlab{a}})Chen, Pang, Wang, Xiong, Li, Sun,
  Feng, Liu, Shi, Ouyang et~al.}]{chen2019hybrid}
Chen, K.; Pang, J.; Wang, J.; Xiong, Y.; Li, X.; Sun, S.; Feng, W.; Liu, Z.;
  Shi, J.; Ouyang, W.; et~al. 2019{\natexlab{a}}.
\newblock Hybrid task cascade for instance segmentation.
\newblock In \emph{CVPR}.

\bibitem[{Chen et~al.(2019{\natexlab{b}})Chen, Wang, Pang, Cao, Xiong, Li, Sun,
  Feng, Liu, Xu, Zhang, Cheng, Zhu, Cheng, Zhao, Li, Lu, Zhu, Wu, Dai, Wang,
  Shi, Ouyang, Loy, and Lin}]{mmdetection}
Chen, K.; Wang, J.; Pang, J.; Cao, Y.; Xiong, Y.; Li, X.; Sun, S.; Feng, W.;
  Liu, Z.; Xu, J.; Zhang, Z.; Cheng, D.; Zhu, C.; Cheng, T.; Zhao, Q.; Li, B.;
  Lu, X.; Zhu, R.; Wu, Y.; Dai, J.; Wang, J.; Shi, J.; Ouyang, W.; Loy, C.~C.;
  and Lin, D. 2019{\natexlab{b}}.
\newblock {MMDetection}: Open MMLab Detection Toolbox and Benchmark.
\newblock \emph{arXiv:1906.07155} .

\bibitem[{Cheng et~al.(2020)Cheng, Wang, Huang, and Liu}]{cheng2020boundary}
Cheng, T.; Wang, X.; Huang, L.; and Liu, W. 2020.
\newblock Boundary-preserving mask R-CNN.
\newblock In \emph{ECCV}.

\bibitem[{Dai et~al.(2016)Dai, He, Li, Ren, and Sun}]{dai2016instance}
Dai, J.; He, K.; Li, Y.; Ren, S.; and Sun, J. 2016.
\newblock Instance-sensitive fully convolutional networks.
\newblock In \emph{ECCV}.

\bibitem[{Dai, He, and Sun(2016)}]{Dai_2016_CVPR}
Dai, J.; He, K.; and Sun, J. 2016.
\newblock Instance-Aware Semantic Segmentation via Multi-Task Network Cascades.
\newblock In \emph{CVPR}.

\bibitem[{Danielczuk et~al.(2019)Danielczuk, Matl, Gupta, Li, Lee, Mahler, and
  Goldberg}]{danielczuk2019segmenting}
Danielczuk, M.; Matl, M.; Gupta, S.; Li, A.; Lee, A.; Mahler, J.; and Goldberg,
  K. 2019.
\newblock Segmenting unknown 3d objects from real depth images using mask r-cnn
  trained on synthetic data.
\newblock In \emph{ICRA}.

\bibitem[{Girshick(2015)}]{girshick2015fast}
Girshick, R. 2015.
\newblock Fast r-cnn.
\newblock In \emph{ICCV}.

\bibitem[{He et~al.(2017)He, Gkioxari, Dollar, and Girshick}]{He_2017_ICCV}
He, K.; Gkioxari, G.; Dollar, P.; and Girshick, R. 2017.
\newblock Mask R-CNN.
\newblock In \emph{ICCV}.

\bibitem[{Homayounfar et~al.(2020)Homayounfar, Xiong, Liang, Ma, and
  Urtasun}]{homayounfar2020levelset}
Homayounfar, N.; Xiong, Y.; Liang, J.; Ma, W.-C.; and Urtasun, R. 2020.
\newblock LevelSet R-CNN: A Deep Variational Method for Instance Segmentation.
\newblock In \emph{ECCV}.

\bibitem[{Kim et~al.(2019)Kim, Ma, Kim, Kim, and Yoo}]{kim2019progressive}
Kim, J.; Ma, M.; Kim, K.; Kim, S.; and Yoo, C.~D. 2019.
\newblock Progressive attention memory network for movie story question
  answering.
\newblock In \emph{CVPR}.

\bibitem[{Kim et~al.(2020)Kim, Ma, Pham, Kim, and Yoo}]{kim2020modality}
Kim, J.; Ma, M.; Pham, T.; Kim, K.; and Yoo, C.~D. 2020.
\newblock Modality Shifting Attention Network for Multi-Modal Video Question
  Answering.
\newblock In \emph{CVPR}.

\bibitem[{Li et~al.(2017)Li, Qi, Dai, Ji, and Wei}]{Li_2017_CVPR}
Li, Y.; Qi, H.; Dai, J.; Ji, X.; and Wei, Y. 2017.
\newblock Fully Convolutional Instance-Aware Semantic Segmentation.
\newblock In \emph{CVPR}.

\bibitem[{Lin et~al.(2017)Lin, Dollar, Girshick, He, Hariharan, and
  Belongie}]{Lin_2017_CVPR}
Lin, T.-Y.; Dollar, P.; Girshick, R.; He, K.; Hariharan, B.; and Belongie, S.
  2017.
\newblock Feature Pyramid Networks for Object Detection.
\newblock In \emph{CVPR}.

\bibitem[{Lin et~al.(2014)Lin, Maire, Belongie, Hays, Perona, Ramanan,
  Doll{\'a}r, and Zitnick}]{lin2014microsoft}
Lin, T.-Y.; Maire, M.; Belongie, S.; Hays, J.; Perona, P.; Ramanan, D.;
  Doll{\'a}r, P.; and Zitnick, C.~L. 2014.
\newblock Microsoft coco: Common objects in context.
\newblock In \emph{ECCV}.

\bibitem[{Liu et~al.(2017)Liu, Jia, Fidler, and Urtasun}]{Liu_2017_ICCV}
Liu, S.; Jia, J.; Fidler, S.; and Urtasun, R. 2017.
\newblock SGN: Sequential Grouping Networks for Instance Segmentation.
\newblock In \emph{ICCV}.

\bibitem[{Liu et~al.(2018)Liu, Qi, Qin, Shi, and Jia}]{Liu_2018_CVPR}
Liu, S.; Qi, L.; Qin, H.; Shi, J.; and Jia, J. 2018.
\newblock Path Aggregation Network for Instance Segmentation.
\newblock In \emph{CVPR}.

\bibitem[{Long, Shelhamer, and Darrell(2015)}]{Long_2015_CVPR}
Long, J.; Shelhamer, E.; and Darrell, T. 2015.
\newblock Fully Convolutional Networks for Semantic Segmentation.
\newblock In \emph{CVPR}.

\bibitem[{Mao et~al.(2018)Mao, Zhang, He, Lin, Kale, Stein, and
  Kostic}]{Mao_2018_CVPR_Workshops}
Mao, T.; Zhang, W.; He, H.; Lin, Y.; Kale, V.; Stein, A.; and Kostic, Z. 2018.
\newblock AIC2018 Report: Traffic Surveillance Research.
\newblock In \emph{CVPRW}.

\bibitem[{Neven et~al.(2018)Neven, De~Brabandere, Georgoulis, Proesmans, and
  Van~Gool}]{neven2018towards}
Neven, D.; De~Brabandere, B.; Georgoulis, S.; Proesmans, M.; and Van~Gool, L.
  2018.
\newblock Towards end-to-end lane detection: an instance segmentation approach.
\newblock In \emph{2018 IEEE Intelligent Vehicles Symposium (IV)}.

\bibitem[{Paszke et~al.(2017)Paszke, Gross, Chintala, Chanan, Yang, DeVito,
  Lin, Desmaison, Antiga, and Lerer}]{paszke2017automatic}
Paszke, A.; Gross, S.; Chintala, S.; Chanan, G.; Yang, E.; DeVito, Z.; Lin, Z.;
  Desmaison, A.; Antiga, L.; and Lerer, A. 2017.
\newblock Automatic differentiation in PyTorch.
\newblock In \emph{NIPS-W}.

\bibitem[{Pathak et~al.(2018)Pathak, Shentu, Chen, Agrawal, Darrell, Levine,
  and Malik}]{pathak2018learning}
Pathak, D.; Shentu, Y.; Chen, D.; Agrawal, P.; Darrell, T.; Levine, S.; and
  Malik, J. 2018.
\newblock Learning instance segmentation by interaction.
\newblock In \emph{CVPRW}.

\bibitem[{Peng et~al.(2017)Peng, Zhang, Yu, Luo, and Sun}]{peng2017large}
Peng, C.; Zhang, X.; Yu, G.; Luo, G.; and Sun, J. 2017.
\newblock Large Kernel Matters--Improve Semantic Segmentation by Global
  Convolutional Network.
\newblock In \emph{CVPR}.

\bibitem[{Pinheiro, Collobert, and Doll{\'a}r(2015)}]{pinheiro2015learning}
Pinheiro, P.~O.; Collobert, R.; and Doll{\'a}r, P. 2015.
\newblock Learning to segment object candidates.
\newblock In \emph{NIPS}.

\bibitem[{Pinheiro et~al.(2016)Pinheiro, Lin, Collobert, and
  Doll{\'a}r}]{pinheiro2016learning}
Pinheiro, P.~O.; Lin, T.-Y.; Collobert, R.; and Doll{\'a}r, P. 2016.
\newblock Learning to refine object segments.
\newblock In \emph{ECCV}.

\bibitem[{Qiao, Chen, and Yuille(2020)}]{qiao2020detectors}
Qiao, S.; Chen, L.-C.; and Yuille, A. 2020.
\newblock DetectoRS: Detecting Objects with Recursive Feature Pyramid and
  Switchable Atrous Convolution.
\newblock \emph{arXiv:2006.02334} .

\bibitem[{Ren et~al.(2015)Ren, He, Girshick, and Sun}]{NIPS2015_5638}
Ren, S.; He, K.; Girshick, R.; and Sun, J. 2015.
\newblock Faster R-CNN: Towards Real-Time Object Detection with Region Proposal
  Networks.
\newblock In \emph{NIPS}.

\bibitem[{Tian, Shen, and Chen(2020)}]{tian2020conditional}
Tian, Z.; Shen, C.; and Chen, H. 2020.
\newblock Conditional Convolutions for Instance Segmentation.
\newblock In \emph{ECCV}.

\bibitem[{Vu et~al.(2019)Vu, Jang, Pham, and Yoo}]{vu2019cascade}
Vu, T.; Jang, H.; Pham, T.~X.; and Yoo, C. 2019.
\newblock Cascade RPN: Delving into High-Quality Region Proposal Network with
  Adaptive Convolution.
\newblock In \emph{NeurIPS}.

\bibitem[{Wang et~al.(2018)Wang, Girshick, Gupta, and He}]{wang2018non}
Wang, X.; Girshick, R.; Gupta, A.; and He, K. 2018.
\newblock Non-local neural networks.
\newblock In \emph{CVPR}.

\bibitem[{Wang et~al.(2020)Wang, Kong, Shen, Jiang, and Li}]{Wang2020Solo}
Wang, X.; Kong, T.; Shen, C.; Jiang, Y.; and Li, L. 2020.
\newblock SOLO: Segmenting Objects by Locations.
\newblock In \emph{ECCV}.

\bibitem[{Xie et~al.(2020)Xie, Sun, Song, Wang, Liu, Liang, Shen, and
  Luo}]{xie2020polarmask}
Xie, E.; Sun, P.; Song, X.; Wang, W.; Liu, X.; Liang, D.; Shen, C.; and Luo, P.
  2020.
\newblock Polarmask: Single shot instance segmentation with polar
  representation.
\newblock In \emph{CVPR}.

\bibitem[{Zhang et~al.(2018)Zhang, Song, Du, and Guizani}]{zhang2018vehicle}
Zhang, Y.; Song, B.; Du, X.; and Guizani, M. 2018.
\newblock Vehicle tracking using surveillance with multimodal data fusion.
\newblock \emph{IEEE Transactions on Intelligent Transportation Systems} .

\bibitem[{Zhang, Fidler, and Urtasun(2016)}]{zhang2016instance}
Zhang, Z.; Fidler, S.; and Urtasun, R. 2016.
\newblock Instance-level segmentation for autonomous driving with deep densely
  connected mrfs.
\newblock In \emph{CVPR}.

\bibitem[{Zhang et~al.(2015)Zhang, Schwing, Fidler, and
  Urtasun}]{Zhang_2015_ICCV}
Zhang, Z.; Schwing, A.~G.; Fidler, S.; and Urtasun, R. 2015.
\newblock Monocular Object Instance Segmentation and Depth Ordering With CNNs.
\newblock In \emph{ICCV}.

\end{thebibliography}

\end{document}